\pdfoutput=1

\documentclass[11pt]{article}

\usepackage{EMNLP2022}

\usepackage{times}
\usepackage{latexsym}

\usepackage[T1]{fontenc}

\usepackage[utf8]{inputenc}

\usepackage{microtype}

\usepackage{inconsolata}

\usepackage{tabularx}
\usepackage{bm}
\usepackage{amsmath}
\usepackage{mathrsfs}
\usepackage{multirow}
\usepackage{booktabs}
\usepackage{tikz}
\usepackage{tikz-qtree}
\usepackage{color}
\usepackage{xcolor}
\usepackage{makecell}
\usepackage[switch]{lineno}
\usepackage{enumitem}
\usepackage{makecell}
\usepackage{bbding}
\usepackage{ulem}
\usepackage{url}

\title{A Fine-grained Interpretability Evaluation Benchmark for Neural NLP}

\author{
Lijie Wang,
Yaozong Shen,
Shuyuan Peng,
Shuai Zhang,
Xinyan Xiao, \\
\textbf{Hao Liu},
\textbf{Hongxuan Tang},
\textbf{Ying Chen},
\textbf{Hua Wu},
\textbf{Haifeng Wang} \\
Baidu Inc, Beijing, China \\
\{wanglijie,shenyaozong@baidu.com\}\\
}

\begin{document}
\maketitle

\begin{abstract}

While there is increasing concern about the interpretability of neural models, the evaluation of interpretability remains an open problem, due to the lack of proper evaluation datasets and metrics. In this paper, we present a novel benchmark to evaluate the interpretability of both neural models and saliency methods. This benchmark covers three representative NLP tasks: sentiment analysis, textual similarity and reading comprehension, each provided with both English and Chinese annotated data. In order to precisely evaluate the interpretability, we provide token-level rationales that are carefully annotated to be sufficient, compact and comprehensive. We also design a new metric, i.e., the consistency between the rationales before and after perturbations, to uniformly evaluate the interpretability on different types of tasks. Based on this benchmark, we conduct experiments on three typical models with three saliency methods, and unveil their strengths and weakness in terms of interpretability. We will release this benchmark\footnote{\url{https://www.luge.ai/\#/luge/task/taskDetail?taskId=15}} and hope it can facilitate the research in building trustworthy systems.


\end{abstract}
\section{Introduction}
\label{sec:intro}

In the last decade, deep learning (DL) has been rapidly developed and has greatly improved various artificial intelligence tasks in terms of accuracy \cite{deng2014deep, litjens2017survey, pouyanfar2018survey}. However, as DL models are black-box systems, their inner decision processes are opaque to users. This lack of transparency makes them untrustworthy and hard to be applied in decision-making applications in fields such as health, commerce and law \cite{fort-couillault-2016-yes}.
Consequently, there is a growing interest in explaining the predictions of DL models \cite{simonyan2014deep, ribeiro2016should, alzantot-etal-2018-generating, bastings-etal-2019-interpretable, jiang-etal-2021-alignment}. Accordingly, many evaluation datasets are constructed and the corresponding metrics are designed to evaluate related works \cite{deyoung-etal-2020-eraser, jacovi-goldberg-2020-towards}.

\begin{table}[tb]
\small
\center
\scalebox{0.95} {
\begin{tabularx}{0.48\textwidth}{ c }
\toprule[1pt]
\textbf{Sentiment Analysis (SA)} \\
\hline
\makecell[l]{\textbf{Instance}$^o$: although it bangs a very cliched drum at times,\\ this crowd-pleaser's \textcolor{red}{fresh} \textcolor{red}{dialogue}, \textcolor{blue}{\uline{energetic} music}, and \\ \textcolor{green}{good-natured} \textcolor{green}{spunk} are often infectious. \\ \textbf{Sentiment label}: positive} \\
\hline
\makecell[l]{\textbf{Instance}$^p$: although it bangs a very cliched drum at times, \\ this crowd-pleaser's \textcolor{red}{novel} \textcolor{red}{dialogue}, \textcolor{blue}{vigorous music}, and \\ \textcolor{green}{good-natured} \textcolor{green}{spunk} are often infectious. \\ \textbf{Sentiment label}: positive} \\
\midrule[1pt]
\textbf{Semantic Textual Similarity (STS)} \\
\hline
\makecell[l]{\textbf{Instance1}$^o$: Is there \uline{a} \textcolor{red}{reason why} we \uline{should} \textcolor{red}{travel alone}? \\ \textbf{Instance2}$^o$: \uline{What are} some \textcolor{red}{reasons to travel alone}? \\ \textbf{Similarity}: same} \\
\hline
\makecell[l]{\textbf{Instance1}$^p$: Is there any \textcolor{red}{reason why} we \textcolor{red}{travel alone}? \\ \textbf{Instance2}$^p$: List some \textcolor{red}{reasons to travel alone}? \\ \textbf{Similarity}: same} \\
\midrule[1pt]
\textbf{Machine Reading Comprehensive (MRC)} \\
\hline
\makecell[l]{\textbf{Question}: \uline{What part of} France were the Normans located? \\ \textbf{Article}$^o$: ...and customs to synthesize a unique ``\textcolor{red}{Norman}'' \\ \textcolor{red}{culture in} the \textcolor{red}{north of} \textcolor{red}{France}. ... \\ \textbf{Answer}: north} \\
\hline
\makecell[l]{\textbf{Question}: Where in France were the Normans located? \\ \textbf{Article}$^p$: ...and customs to synthesize a unique ``\textcolor{red}{Norman}'' \\ \textcolor{red}{culture in} the \textcolor{red}{north of} \textcolor{red}{France}. ... \\ \textbf{Answer}: north} \\
\bottomrule[1pt]
\end{tabularx}
}
\caption{Examples from our benchmark. In each instance, colored tokens are rationales, and tokens in the same color constitute an independent rationale set. Each perturbed example ($^p$) is created on an original example ($^o$), where underlined tokens in the original example have been altered. The consistency of rationales under perturbations is used to evaluate interpretability.}
\label{tab:case_intro}
\end{table}

In order to accurately evaluate model interpretability\footnote{Despite fine-grained distinctions between ``interpretability'' and ``explainability'', we use them interchangeably.} with human-annotated rationales\footnote{In this paper, we focus on highlight-based rationales, which consist of input elements, such as words and sentences, that play a decisive role in the model prediction.} (i.e., evidence that supports the model prediction), many researchers successively propose the properties that a rationale should satisfy, e.g., sufficiency, compactness and comprehensiveness (see Section \ref{ssec:rationale-annotation} for their specific definitions) \cite{kass1988need, fischer1990minimalist, lei-etal-2016-rationalizing, yu-etal-2019-rethinking}. 
However, the existing datasets are designed for different research aims with different metrics, and their rationales do not satisfy all properties needed, as shown in Table \ref{tab:exist_data}, which makes it difficult to track and facilitate the research progress of interpretability.
In addition, all existing datasets are in English.

Meanwhile, many studies focus on designing guidelines and metrics for interpretability evaluation, where plausibility and faithfulness are proposed to measure interpretability from different perspectives \cite{herman2017promise, alvarez2018towards, yang2019evaluating, wiegreffe-pinter-2019-attention, jacovi-goldberg-2020-towards}. 
Plausibility measures how well the rationales provided by models align with human-annotated rationales. With different annotation granularities, token-level and span-level F1-scores are proposed to measure plausibility \cite{deyoung-etal-2020-eraser, mathew2021hatexplain}. 
Faithfulness measures to what extent the provided rationales influence the corresponding predictions. Some studies \cite{yu-etal-2019-rethinking, deyoung-etal-2020-eraser} propose to compare the model's prediction on the full input to its prediction on input masked according to the rationale and its complement (i.e., non-rationale).
However, it is difficult to apply this evaluation method to non-classification tasks, such as machine reading comprehension.
Furthermore, the model prediction on the non-rationale has gone beyond the standard output scope, e.g., the prediction label on the non-rationale should be neither positive nor negative in the sentiment classification task. 
Thus the metric provided by this method can not generally and may not precisely evaluate the interpretability.

In order to address the above problems, we release a new interpretability evaluation benchmark which provides fine-grained rationales for three tasks and a new evaluation metric for interpretability. 
Our contributions include: 
\begin{itemize}[leftmargin=*]
\item Our benchmark contains three representative tasks in both English and Chinese, i.e., sentiment analysis, semantic textual similarity and machine reading comprehension. Importantly, all annotated rationales meet the requirements of sufficiency, compactness and comprehensiveness by being organized in the set form.
\item To precisely and uniformly evaluate the interpretability of all tasks, we propose a new evaluation metric, i.e., the consistency between the rationales provided on examples before and after perturbation. The perturbations are crafted in a way that will not change the model decision mechanism. This metric measures model fidelity under perturbations and could help to find the relationship between interpretability and other metrics, such as robustness.
\item We give an in-depth analysis based on three typical models with three popular saliency methods, as well as a comparison between our proposed metrics and the existing metrics. The results show that our benchmark can be used to evaluate the interpretability of DL models and saliency methods.
Meanwhile, the results strongly indicate that the research on interpretability of NLP models has much further to go, and we hope our benchmark will do its bit along the way.
\end{itemize}

\begin{table*}[tb]
\small
\renewcommand\tabcolsep{2.5pt}
\centering
\scalebox{0.95} {
\begin{tabular}{l c c c c}
\toprule
\multirow{2}{*}{Datasets} & \multirow{2}{*}{Granularity} & \multicolumn{3}{c}{Properties}\\
\cline{3-5}
 & & Sufficiency & Compactness & Comprehensiveness\\
\hline
e-SNLI$^\star$ \cite{camburu2018snli} & word & \XSolidBrush & \Checkmark & \XSolidBrush \\
HUMMINGBIRD \cite{hayati-etal-2021-bert} & word & \Checkmark$^-$ & \XSolidBrush & -- \\
HateXplain \cite{mathew2021hatexplain} & word & \Checkmark$^-$ & -- & \Checkmark \\
\hline
Movie Reviews$^\star$ \cite{zaidan-eisner-2008-modeling} & snippet & \Checkmark & \XSolidBrush & \XSolidBrush \\
CoS-E$^\star$ \cite{rajani-etal-2019-explain} & snippet & \Checkmark$^-$ & \XSolidBrush & \Checkmark \\
Evidence Inference$^\star$ \cite{lehman-etal-2019-inferring} & snippet & \Checkmark & \XSolidBrush & \XSolidBrush \\
BoolQ$^\star$ \cite{deyoung-etal-2020-eraser} & snippet &  \Checkmark & \XSolidBrush & \Checkmark \\
\hline
WikiQA \cite{yang-etal-2015-wikiqa} & sentence & \Checkmark & \XSolidBrush & -- \\
MultiRC$^\star$ \cite{khashabi-etal-2018-looking} & sentence & \Checkmark & \XSolidBrush & \Checkmark \\
HotpotQA \cite{yang-etal-2018-hotpotqa} & sentence & \Checkmark & \XSolidBrush & \Checkmark \\
FEVER$^\star$ \cite{thorne-etal-2018-fever} & sentence & \Checkmark & \XSolidBrush & -- \\
SciFact \cite{wadden-etal-2020-fact} & sentence & \Checkmark & \XSolidBrush & -- \\
\hline
Ours & word & \Checkmark & \Checkmark & \Checkmark \\
\bottomrule
\end{tabular}
}
\caption{Statistics of existing datasets with highlight-based rationales. The datasets marked with $^\star$ are collected and modified by ERASER \cite{deyoung-etal-2020-eraser}. ERASER manually reviews and constructs snippet-level rationales to make them satisfy sufficiency and comprehensiveness. \Checkmark$^-$ represents the rationale contains key words, but does not contain enough information for the prediction. The value `-' represents the property is not mentioned in the paper.}
\label{tab:exist_data}
\end{table*}

\section{Related Work}
\label{sec:related_work}

As our work provides a new interpretability evaluation benchmark with human-annotated rationales, in this section, we mainly introduce saliency methods for the rationale extraction, interpretability evaluation datasets and metrics.

\paragraph{Saliency Methods} In the post-hoc interpretation research field, saliency methods are widely used to interpret model decisions by assigning a distribution of importance scores over the input tokens to represent their impacts on model predictions \cite{simonyan2014deep, ribeiro2016should, murdoch2018beyond}. 
They are mainly divided into four categories: gradient-based, attention-based, erasure-based and linear-based. In gradient-based methods, the magnitudes of the gradients serve as token importance scores \cite{simonyan2014deep, smilkov2017smoothgrad, sundararajan2017axiomatic}.
Attention-based methods use attention weights as token importance scores \cite{jain-wallace-2019-attention, wiegreffe-pinter-2019-attention}. 
In erasure-based methods, the token importance score is measured by the change of output when the token is removed \cite{li2016understanding, feng-etal-2018-pathologies}.
Linear-based methods use a simple and explainable linear model to approximate the evaluated model behavior locally and use the learned token weights as importance scores \cite{ribeiro2016should, alvarez-melis-jaakkola-2017-causal}. 
These methods have their own advantages and limitations from aspects of computational efficiency, interpretability performance and so on \cite{nie2018theoretical, jain-wallace-2019-attention, de-cao-etal-2020-decisions, sixt2020explanations}. 

\paragraph{Interpretability Datasets} 
Many datasets with human-annotated rationales have been published for interpretability evaluation, e.g., highlight-based rationales \cite{deyoung-etal-2020-eraser, mathew2021hatexplain}, free-text rationales \cite{camburu2018snli, rajani-etal-2019-explain} and structured rationales \cite{ye-etal-2020-teaching, geva-etal-2021-aristotle}. 
To create high-quality highlight-based rationales, many studies give their views on the properties that a rationale should satisfy. 
\citet{kass1988need} propose that a rationale should be understood by humans.
\citet{lei-etal-2016-rationalizing} point that a rationale should be compact and sufficient, i.e., it is short and contains enough information for a prediction.
\citet{yu-etal-2019-rethinking} introduce comprehensiveness as a criterion, requiring all rationales to be selected, not just a sufficient set.
Although the above criteria have been proposed for highlight rationales, the existing datasets in Table \ref{tab:exist_data} are built with part of them, as they are conducted on different tasks with individual aims.

\paragraph{Interpretability Metrics} 
For highlight-based rationales, plausibility and faithfulness are often used to measure interpretability from the aspects of human cognition and model fidelity \cite{arras2017relevant, mohseni2018human, weerts2019human}.
\citet{deyoung-etal-2020-eraser} propose to use IOU (Intersection-Over-Union) F1-score and AUPRC (Area Under the Precision-Recall curve) score to measure plausibility of snippet-level rationales.
\citet{mathew2021hatexplain} use token F1-score to evaluate plausibility of token-level rationales.
\citet{jacovi-goldberg-2020-towards} provide concrete guidelines for the definition and evaluation of faithfulness.
\citet{deyoung-etal-2020-eraser} propose to evaluate faithfulness from the perspectives of sufficiency and comprehensiveness of rationales (Equation \ref{equation:eraser-faithful}). 
However, this evaluation manner is only applicable to classification tasks and brings uncontrollable factors to interpretability evaluation.
Thus \citet{yin-etal-2022-sensitivity} propose sensitivity and stability as complementary metrics for faithfulness.
\citet{ding-koehn-2021-evaluating} evaluate faithfulness of saliency methods on natural language models by measuring how consistent the rationales are regarding perturbations.

In this work, we provide a new interpretability evaluation benchmark, containing fine-grained annotated rationales, a new evaluation metric and the corresponding perturbed examples.

\section{Evaluation Data Construction}
\label{sec:dataset}

\begin{figure}[tb]
\centering
\includegraphics[width=0.45\textwidth]{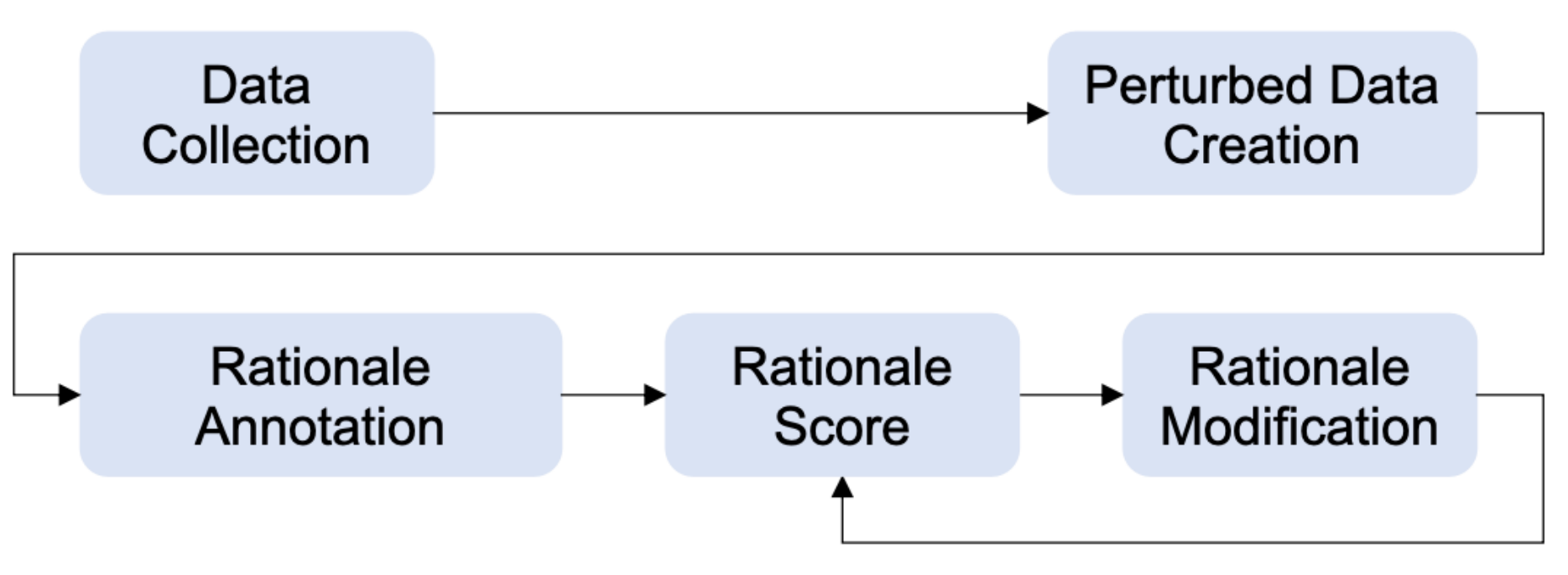}
\caption{The construction workflow of our datasets.}
\label{fig:data-workflow}
\end{figure}

As illustrated in Figure \ref{fig:data-workflow}, the construction of our datasets mainly consists of three steps: 1) data collection for each task; 2) perturbed data construction; 3) iterative rationale annotation and checking. 
We first introduce the annotation process, including the annotation criteria for perturbations and rationales. Then we describe our data statistics. In addition, we show other annotation details in Appendix \ref{sec:data_other}.

\subsection{Data Collection}
\label{ssec:data-collection}

In order to provide a general and unified interpretability evaluation benchmark, we construct evaluation datasets for three representative tasks, i.e., sentiment analysis, semantic textual similarity, and machine reading comprehension. Meanwhile, we create both English and Chinese evaluation datasets for each task. 

\textbf{Sentiment Analysis (SA)}, a single-sentence classification task, aims to predict a sentiment label for the given instance. For English, we randomly select 1,500 instances from Stanford Sentiment Treebank (SST) \cite{socher-etal-2013-recursive} dev/test sets, and 400 instances from Movie Reviews \cite{zaidan-eisner-2008-modeling} test set. For Chinese, we randomly sample 60,000 instances from the logs of an open SA API\footnote{\url{https://ai.baidu.com/tech/nlp_apply/sentiment_classify}. Due to the diversity of these logs, we choose instances from these logs for annotation.} with the permission of users. The annotators select instances for annotation (see Appendix \ref{sec:data_other} for details) and label a sentiment polarity for each unlabeled instance. Then 2,000 labeled instances are chosen for building evaluation set.

\textbf{Semantic Textual Similarity (STS)}, a sentence-pair similarity task, is to predict the similarity between two instances. We randomly select 2,000 pairs from Quora Question Pairs (QQP) \cite{wang-etal-2018-glue} and LCQMC \cite{liu-etal-2018-lcqmc} to build English and Chinese evaluation data respectively.

\textbf{Machine Reading Comprehension (MRC)}, a long-text comprehension task, aims to extract an answer based on the question and the corresponding passage. We randomly select 1,500 triples with answers and 500 triples without answers from SQUAD2.0 \cite{rajpurkar-etal-2018-know} and DuReader \cite{he-etal-2018-dureader} for building English and Chinese evaluation set respectively. 

\subsection{Perturbed Data Creation}
\label{ssec:perturbed-creation}

Recent studies \cite{jacovi-goldberg-2020-towards, ding-koehn-2021-evaluating} claim that a saliency method is faithful if it provides similar rationales for similar inputs and outputs.
Inspired by them, we propose to evaluate the model faithfulness via measuring how consistent its rationales are regarding perturbations that are supposed to preserve the same model decision mechanism.
In other words, under perturbations, a model is considered to be faithful if the change of its rationales is consistent with the change of its prediction.
Consequently, we construct perturbed examples for each original input.

\paragraph{Perturbation Criteria} 
Perturbations should not change the model internal decision mechanism.
We create perturbed examples from two aspects: 1) perturbations do not influence model rationales and predictions; 2) perturbations cause the alterations of rationales and may change predictions. Please note that the influence of perturbations comes from human's basic intuition on model's decision-making mechanism. Based on the literature \cite{jia-liang-2017-adversarial, mccoy-etal-2019-right, ribeiro-etal-2020-beyond}, we define three perturbation types.
\begin{itemize}[leftmargin=*]
\item \textbf{Alteration of dispensable words}. Insert, delete and replace words that should have no effect on model predictions and rationales, e.g., the sentence ``\textit{what are} some reasons to travel alone'' is changed to ``\textit{list} some reasons to travel alone''.
\item \textbf{Alteration of important words}. Replace important words which have an impact on model predictions with their synonyms or related words, such as ``i dislike you'' instead of ``i hate you''. In this situation, the model prediction and rationale should change with perturbations. 
\item \textbf{Syntax transformation}. Transform the syntax structure of an instance without changing its semantics, e.g., ``the customer commented the hotel'' is transformed into ``the hotel is commented by the customer''. In this case, the model prediction and rationale should not be affected. 
\end{itemize}

For each original input, the annotator first selects a perturbation type, then creates a perturbed example according to the definition of this perturbation type. Please note that the annotators can select more than one perturbation type for an original input. 
We ask the annotator to create at least one perturbed example for each original input. And they need to create at least 100 perturbed examples for each perturbation type. For each task, we have two annotators to create perturbed examples and label golden results for these examples, i.e., sentiment label for SA, similarity label for STS and answer for MRC.
According to the perturbation criteria, most of the perturbed examples have the same results as their original ones.
Then we ask the other two annotators to review and modify the created examples and their corresponding results. Since the annotation task in this step is relatively easy, the accuracy of created examples after checking is more than 95\%.

\subsection{Iterative Rationale Annotation Process}
\label{ssec:rationale-annotation}

Given an input and the corresponding golden result, the annotators highlight important input tokens that support the prediction of golden result as the rationale. Then we introduce the rationale criteria and the annotation process used in our work.

\paragraph{Rationale Criteria} 
As discussed in recent studies \cite{lei-etal-2016-rationalizing, yu-etal-2019-rethinking}, a rationale should satisfy the following properties.
\begin{itemize}[leftmargin=*]
\item \textbf{Sufficiency}. A rationale is sufficient if it contains enough information for people to make the correct prediction. In other words, people can make the correct prediction only based on tokens in the rationale.
\item \textbf{Compactness}. A rationale is compact if all of its tokens are indeed required in making a correct prediction. That is to say, when any token is removed from the rationale, the prediction will change or become difficult to make.
\item \textbf{Comprehensiveness}. A rationale is comprehensive if its complements in the input can not imply the prediction, that is, all evidence that supports the output should be labeled as rationales.
\end{itemize}

\paragraph{Annotation Process}
To ensure the data quality, we adopt an iterative annotation workflow, consisting of three steps, as described in Figure \ref{fig:data-workflow}.

\textbf{Step 1: rationale annotation}. Based on human's intuitions on the model decision mechanism, given the input and the corresponding golden result, the ordinary annotators who are college students majoring in languages label all critical tokens to guarantee the rationale's comprehensiveness. Then they organize these tokens into several sets, each of which should be sufficient and compact. That is to say, each set can support the prediction independently. As described in Table \ref{tab:case_intro}, the first example contains three rationale sets, and tokens in the same color belong to the same set. Based on this set form, the rationale satisfies the above three criteria.

\textbf{Step 2: rationale scoring}. Our senior annotators\footnote{They are full-time employees, and have lots of experience in annotating data for NLP tasks.} double-check the annotations by scoring the given rationales according to the annotation criteria. For each rationale set, the annotators rate their confidences for sufficiency and compactness.
The confidences for \textbf{sufficiency} consist of three classes: \textit{can not support result (1)}, \textit{not sure (2)} and \textit{can support result (3)}. 
And the confidences for \textbf{compactness} compose of four classes: \textit{include redundant tokens (1)}, \textit{include disturbances (2)}, \textit{not sure (3)} and \textit{conciseness (4)}. 
Then based on all rationale sets for each input, the annotators rate their confidences for \textbf{comprehensiveness} on a 3-point scale including \textit{not be comprehensive (1)}, \textit{not sure (2)}, \textit{be comprehensive (3)}.

A rationale is considered to be of high-quality if its average score on sufficiency, compactness and comprehensiveness is equal to or greater than $3.0$, $3.6$, $2.6$. That is to say, at least two-thirds of the annotators give the highest confidence, and less than one-third of the annotators give the confidence of ``\textit{not sure}''. Then all unqualified data whose average score on a property is lower than the corresponding threshold goes to the next step.

\textbf{Step 3: rationale modification}. Low-quality rationales are shown to the ordinary annotators again. The annotators correct the rationales to meet the properties with scores below the threshold. 

Then the corrected rationales are scored by senior annotators again. The unqualified data after three loops is discarded. This iterative annotation-scoring process can ensure the data quality.  

Other annotation details, such as annotator information, annotation training and data usage instructions, are described in Appendix \ref{sec:data_other}.

\begin{table}[tb]
\small
\center
\scalebox{0.9}{
\begin{tabular}{l| c c c | c c c}
\toprule
\multirow{2}{*}{Tasks} & \multicolumn{3}{c|}{\textbf{English}} & \multicolumn{3}{c}{\textbf{Chinese}} \\
\cline{2-7}
 & Size & RLR & RSN & Size & RLR & RSN\\
\hline
SA & 1,999 & 20.1\% & 2.1 & 2,160 & 27.6\% & 1.4\\
STS & 2,248 & 50.4\% & 1.0 & 2,146 & 66.6\% & 1.0\\
MRC & 1,969 & 10.4\% & 1.0 & 2,315 & 9.8\% & 1.0\\
\bottomrule
\end{tabular}
}
\caption{Overview of our datasets. ``Size'' shows the number of original/perturbed pairs. ``RLR'' represents the ratio of rationale length to its input length. ``RSN'' represents the number of rationale sets in an input. We report the average RLR and RSN over all data.}
\label{tab:data_stas}
\end{table} 

\subsection{Data Statistics}
\label{ssec:data_stas}
We give a comparison between our benchmark and other existing datasets, as shown in Table \ref{tab:exist_data}. Compared with existing datasets, our benchmark contains three NLP tasks with both English and Chinese annotated data. Compared with ERASER which collects seven existing English datasets in its benchmark and provides snippet-level rationales to satisfy sufficiency and comprehensiveness, our benchmark provides token-level rationales and satisfies all three primary properties of rationales.  

Table \ref{tab:data_stas} shows the detailed statistics of our benchmark. We can see that the length ratio and the number of rationales vary with datasets and tasks, where the length ratio affects the interpretability performance on plausibility, as shown in Table \ref{tab:eval_inter_all}. 

Meanwhile, we evaluate the sufficiency of human-annotated rationales by evaluating model performance on rationales, as shown in Table \ref{tab:rationale_eval}. Despite the input construction based on rationales has destroyed the distribution of original inputs, model performance on human-annotated rationales is competitive with that on full inputs, especially on MRC task and Chinese datasets. We can conclude that human-annotated rationales are sufficient. Meanwhile, we give more data analysis in Table \ref{tab:eval_rationale_more}, such as model performance on non-rationales, sufficiency and comprehensiveness scores. 

\begin{table}[tb]
\small
\center
\scalebox{0.83} {
\begin{tabular}{l | cc | cc | cc}
\toprule
\multirow{2}{*}{Models} & \multicolumn{2}{c|}{\textbf{SA}} & \multicolumn{2}{c}{\textbf{STS}} & \multicolumn{2}{c}{MRC} \\
 & Acc$^f$ & Acc$^r$ & Acc$^f$ & Acc$^r$ & F1$^{f}$ & F1$^{r}$\\
\hline
\multicolumn{7}{c}{\textbf{English} } \\
\cline{1-7}
LSTM & 78.2 & \textbf{86.2} & 74.6 & 69.8 & 54.4 & 53.4 \\
RoBERTa-base & 93.8 & 92.4 & 92.7 & 89.3 & 71.7 & \textbf{80.8} \\
RoBERTa-large & 95.4 & 91.5 & 93.2 & 88.8 & 76.0 & \textbf{76.7} \\
\hline
\multicolumn{7}{c}{\textbf{Chinese}} \\
\cline{1-7}
LSTM & 60.0 & \textbf{70.4} & 75.2 & \textbf{80.7} & 66.4 & \textbf{82.2} \\
RoBERTa-base & 59.8 & \textbf{77.0} & 85.5 & \textbf{88.1} & 65.8 & \textbf{89.3} \\
RoBERTa-large & 62.6 & \textbf{80.6} & 86.0 & \textbf{87.4} & 67.8 & \textbf{83.3} \\
\bottomrule
\end{tabular}
} 
\caption{Model performance on the original full input (Acc$^f$) and  human-annotated rationale (Acc$^r$).}
\label{tab:rationale_eval}
\end{table}

\section{Metrics}
\label{sec:metric}
Following existing studies \cite{deyoung-etal-2020-eraser, ding-koehn-2021-evaluating, mathew2021hatexplain}, we evaluate interpretability from the perspectives of plausibility and faithfulness. Plausibility measures how well the rationales provided by the model agree with human-annotated ones. And faithfulness measures the degree to which the provided rationales influence the corresponding predictions.

Different from existing work, we adopt \textbf{token-F1} score for plausibility and propose a new metric \textbf{MAP} for faithfulness.


\textbf{Token F1-score} is defined in Equation \ref{equation:marco-f1}, which is computed by overlapped rationale tokens. Since an instance may contain multiple golden rationale sets, for the sake of fairness, we take the set that has the largest F1-score with the predicted rationale as the ground truth for the current prediction. 
\begin{equation}
\small
\centering
\begin{aligned}
\texttt{Token-F1} &= \frac{1}{N}\sum_{i=1}^N (2 \times \frac{P_i \times R_i}{P_i+R_i}) \\
\texttt{where} \quad P_i&=\frac{|S_i^p \cap S_i^g|}{|S_i^p|} \,\, \texttt{and} \,\, R_i=\frac{|S_i^p \cap S_i^g|}{|S_i^g|}
\end{aligned}
\label{equation:marco-f1}
\end{equation}
where $S_i^p$ and $S_i^g$ represent the rationale set of $i$-th instance provided by models and human respectively; $N$ is the number of instances.

\textbf{MAP} (Mean Average Precision) measures the consistency of rationales under perturbations and is used to evaluate faithfulness. According to the original/perturbed input pair, MAP aims to calculate the consistency of two token lists sorted by token importance score, as defined in Equation \ref{equation:map}. The high MAP indicates the high consistency.
\begin{equation}
\small
\centering
\texttt{MAP} = \frac{\sum_{i=1}^{|X^p|}(\sum_{j=1}^i G(x_j^p, X_{1:i}^o))/i}{|X^p|}
\label{equation:map}
\end{equation}
where $X^o$ and $X^p$ represent the sorted rationale token list of the original and perturbed inputs, according to the token important scores assigned by a specific saliency method. $|X^p|$ represents the number of tokens in $X^p$. $X_{1:i}^o$ consists of top-$i$ important tokens of $X^o$. The function $G(x,Y)$ is to determine whether the token $x$ belongs to the list $Y$, where $G(x,Y)=1 \, \texttt{iff} \, x \in Y$. 

Meanwhile, we also report results of metrics proposed in  \citet{deyoung-etal-2020-eraser}, i.e., IOU F1-score for plausibility, and the joint of sufficiency and comprehensiveness for faithfulness.

\textbf{IOU F1-score} is proposed on span-level rationales, which is the size of token overlap in two sets divided by the size of their union, as shown by $S_i$ in Equation \ref{equation:iou-f1}. A rationale is considered as a match if its $S_i$ is equal to or greater than $0.5$, as illustrated by the $Greater$ function.
\begin{equation}
\small
\centering
\begin{aligned}
\texttt{IOU-F1} &= \frac{1}{N}\sum_{i=1}^N Greater(S_i, 0.5) \\
\texttt{where} \quad S_i &= \frac{|S_i^p \cap S_i^g|}{|S_i^p \cup S_i^g|} 
\end{aligned}
\label{equation:iou-f1}
\end{equation}

The joint of \textbf{sufficiency} ($\texttt{Score-Suf}$) and \textbf{comprehensiveness} ($\texttt{Score-Com}$) is shown in Equation \ref{equation:eraser-faithful}. A lower sufficiency score implies the rationale is more sufficient and a higher comprehensiveness score means the rationale is more influential in the prediction. A faithful rationale should have a low sufficiency score and a high comprehensiveness score.
\begin{equation}
\small
\centering
\begin{aligned}
\texttt{Score-Suf} &= \frac{1}{N}\sum_{i=1}^N (F(x_i)_j - F(r_i)_j) \\
\texttt{Score-Com} &= \frac{1}{N}\sum_{i=1}^N (F(x_i)_j - F(x_i \setminus r_i)_j)
\end{aligned}
\label{equation:eraser-faithful}
\end{equation}
where $F(x_i)_j$ represents the prediction probability provided by the model $F$ for class $j$ on the input $x_i$; $r_i$ represents the rationale of $x_i$, and $x_i\setminus{r_i}$ represents its non-rationale.

\section{Experiments}
\label{sec:experiment}

\subsection{Experiment Settings}
\label{sec:exp_set}
We implement three widely-used models and three saliency methods. We give brief descriptions of them and leave the implementation details to Appendix \ref{sec:att-extraction}. The source code will be released with our evaluation datasets.

\paragraph{Saliency Methods}
We adopt integrated gradient (\textbf{IG}) method \cite{sundararajan2017axiomatic}, attention-based (\textbf{ATT}) method \cite{jain-wallace-2019-attention} and linear-based (\textbf{LIME}) \cite{ribeiro2016should} method in our experiments. 
IG assigns importance score for each token by integrating the gradient along the path from a defined input baseline to the original input. 
ATT uses attention weights as importance scores, and the acquisition of attention weights depends on the specific model architecture. 
LIME uses the token weights learned by the linear model as importance scores. 

For each saliency method, we take the top-$k^d$ important tokens to compose the rationale for an input, where $k^d$ is the product of the current input length and the average rationale length ratio of a dataset $d$, as shown by \textit{RLR} in Table \ref{tab:data_stas}. 

\begin{table}[tb]
\small
\center
\scalebox{0.88} {
\begin{tabular}{l | cc | cc | cc}
\toprule
\multirow{2}{*}{Models} & \multicolumn{2}{c|}{\textbf{SA} (Acc)} & \multicolumn{2}{c|}{\textbf{STS} (Acc)} & \multicolumn{2}{c}{\textbf{MRC} (F1)} \\
 & Ori & Ours & Ori & Ours & Ori & Ours \\
\hline
\multicolumn{7}{c}{\textbf{English} } \\
\cline{1-7}
LSTM & 78.6 & 78.2 & 78.6 & 74.6 & 58.6 & 54.4 \\
RoBERTa-base & 92.1 & 93.8 & 91.5 & 92.7 & 78.4 & 71.7 \\
RoBERTa-large & 91.3 & 95.4 & 91.4 & 93.2 & 83.8 & 76.0 \\
\hline
\multicolumn{7}{c}{\textbf{Chinese}} \\
\cline{1-7}
LSTM & 86.7 & 60.0 & 77.4 & 75.2 & 75.0 & 66.4 \\
RoBERTa-base & 95.1 & 59.8 & 88.1 & 85.5 & 74.4 & 65.8 \\
RoBERTa-large & 95.0 & 62.6 & 88.1 & 86.0 & 77.8 & 67.8 \\
\bottomrule
\end{tabular}
} 
\caption{Conventional performance of base models on three tasks, where ``Acc'' is short for accuracy. The ``Ori'' dev/test set comes from the same dataset as training set. ``Ours'' represents our evaluation datasets.}
\label{tab:eval_acc_all}
\end{table} 

\paragraph{Comparison Models} For each task, we re-implement three typical models with different network architectures and parameter sizes, namely LSTM \cite{hochreiter1997long}, RoBERTa-base and RoBERTa-large \cite{liu2019roberta}. Based on these backbone models, we then fine-tune them with commonly-used datasets of three specific tasks. 
For SA, we select training sets of SST and ChnSentiCorp\footnote{\tiny{\url{https://github.com/pengming617/bert_classification}}} to train models for English and Chinese respectively. For STS, training sets of QQP and LCQMC are used to train English and Chinese models. For MRC, SQUAD2.0 and DuReader are used as training sets for English and Chinese respectively. For each task, we select the best model on the original dev set. 

In order to confirm the correctness of our implementation, Table \ref{tab:eval_acc_all} shows model performances on both original dev/test and our evaluation datasets. We can see that our re-implemented models output close results reported in related works \cite{liu-etal-2018-lcqmc, wang2017machine, liu2019roberta}. 
Meanwhile, the results of Chinese SA and MRC tasks decrease significantly on our evaluation sets. This may be caused by the poor generalization and robustness of the model, as our evaluation datasets contain perturbed examples and Chinese data for SA is not from the ChnSentiCorp dataset.

\begin{table*}[tb]
\renewcommand\tabcolsep{2.5pt}
\small
\centering
\scalebox{0.83} {
\begin{tabular}{l | c c c c c | c c c c c | c c c}
\toprule
\multirow{3}{*}{Models + Methods} & \multicolumn{5}{c|}{SA} & \multicolumn{5}{c|}{STS} & \multicolumn{3}{c}{MRC}\\
\cline{2-6} \cline{7-11} \cline{12-14}
 & \multicolumn{2}{c|}{Plausibility} & \multicolumn{3}{c|}{Faithfulness} & \multicolumn{2}{c|}{Plausibility} & \multicolumn{3}{c|}{Faithfulness} &
 \multicolumn{2}{c|}{Plausibility} & 
 \multicolumn{1}{c}{Faithfulness} \\
\cline{2-3} \cline{4-6} \cline{7-8} \cline{9-11} \cline{12-13} \cline{14-14} 
 & Token-F1$\uparrow$ & IOU-F1$\uparrow$ & MAP$\uparrow$ & Suf$\downarrow$ & Com$\uparrow$ & Token-F1 & IOU-F1 & MAP & Suf & Com & Token-F1 & IOU-F1 & MAP \\
\hline
LSTM + IG & 36.9 & 12.1 & 67.2 & -0.025 & 0.708 & 54.1 & 17.3 & 69.0 & 0.048 & 0.441 & 40.7 & 11.0 & 72.3 \\
RoBERTa-base + IG & 37.4 & 10.4 & 64.1 & 0.059 & 0.392 & 52.9 & 24.2 & 65.3 & 0.153 & 0.478 & 42.1 & 11.0 & 66.9 \\
RoBERTa-large + IG & 35.0 & 7.9 & 40.6 & 0.130 & 0.260 & 52.7 & 35.9 & 49.7 & 0.224 & 0.400 & 18.0 & 0.1 & 18.0 \\
\hline
LSTM + ATT & 36.6 & 12.4 & 67.8 & 0.123 & 0.298 & 49.6 & 11.8 & 76.0 & 0.221 & 0.313 & 19.9 & 0.4 & 88.3 \\
RoBERTa-base + ATT & 33.2 & 9.4 & 69.2 & 0.267 & 0.128 & 66.5 & 54.2 & 73.6 & 0.185 & 0.337 & 22.6 & 2.6 & 55.0 \\
RoBERTa-large + ATT & 23.3 & 3.1 & 75.9 & 0.301 & 0.095 & 56.8 & 35.9 & 75.4 & 0.136 & 0.399 & 26.6 & 1.3 & 76.0 \\
\hline
LSTM + LIME & 36.6 & 11.3 & 63.2 & -0.040 & 0.762 & 54.5 & 19.2 & 60.0 & 0.134 & 0.311 & - & - & - \\
RoBERTa-base + LIME & 41.5 & 13.8 & 61.0 & 0.032 & 0.568 & 58.7 & 34.9 & 70.5 & 0.064 & 0.509 & - & - & - \\
RoBERTa-large + LIME & 41.4 & 14.3 & 62.9 & 0.053 & 0.505 & 61.2 & 42.3 & 71.8 & 0.019 & 0.524 & - & - & - \\
\bottomrule
\end{tabular}
}
\caption{Interpretability evaluation results on English datasets of three tasks. The metric with $\uparrow$ means the higher the score, the better the performance. Conversely, $\downarrow$ means a low score represents a good performance. As LIME is specially designed for classification tasks, we have not applied it to MRC. Meanwhile, the sufficiency score ($\texttt{Suf}$) and the comprehensiveness score ($\texttt{Com}$) are also only suitable for classification tasks, as shown in Equation \ref{equation:eraser-faithful}. Thus we do not report these two scores on MRC.}
\label{tab:eval_inter_all}
\end{table*} 


\subsection{Evaluation Results}
\label{ssec:evaluation_result}
Table \ref{tab:eval_inter_all} shows the evaluation results of interpretability from the plausibility and faithfulness perspectives. 
Within the scope of baseline models and saliency methods used in our experiments, there are three main findings. 
First, based on all models and saliency methods used in our experiments, \emph{our metrics for interpretability evaluation, namely token-F1 score and MAP, are more fine and generic}, especially MAP, which applies to all three tasks. 
Second, \emph{IG method performs better on plausibility and ATT method performs better on faithfulness}. Meanwhile, ATT method achieves best performance in sentence-pair tasks.
Third, with all three saliency methods, in these three tasks, \emph{LSTM model is comparable with transformer model (i.e., RoBERTa based model in our experiments) on interpretability}, though LSTM performs worse than transformer in term of accuracy. We think that the generalization ability of LSTM model is weak, leading to low accuracy, even with relatively reasonable rationales.

In the following paragraphs, we first give a comparison between our proposed metrics and those used in related studies. Then we give a detailed analysis about the interpretability results of three saliency methods and three evaluated models.

\paragraph{Comparison between Evaluation Metrics}
We report results of token-F1 and IOU-F1 scores for plausibility. The higher the scores, the more plausible the rationales. 
It can be seen that the two metrics have the similar trends in all three tasks with all three saliency methods.
But token-F1 is much precise than IOU-F1, as the IOU-F1 score of a rationale is 1 only if its overlap with ground truth is no less than 0.5 (Equation \ref{equation:iou-f1}).
However, in all three tasks, overlaps of most instances are less than 0.5, especially in the task with a low \textit{RLR}.
Thus IOU-F1 is too coarse to evaluate token-level rationales.
Instead, token-F1 focuses on evaluating token impact on model predictions, so as to be more suitable for evaluating compact rationales.

For faithfulness evaluation, we report results of MAP, sufficiency and comprehensiveness scores.
We can see that our proposed MAP is an efficient metric for faithfulness evaluation.
Specifically, it applies to most tasks, especially non-classification tasks. 
Moreover, in the two classification tasks (i.e., SA and STS), with IG and LIME methods, MAP has the same trend as the other two metrics over all three models, which further indicates that MAP can well evaluate the faithfulness of rationales. 
With ATT method, there is no consistent relationship between these three metrics. 
We think this is because the calculations of sufficiency and comprehensiveness scores with ATT method are not accurate and consistent enough. 
For example, in the SA task, from the comparison of three saliency methods with LSTM model, we can see that the rationales extracted by these methods have similar plausibility scores, but the sufficiency score with ATT method is much higher than that with the other two methods. Please note that a low sufficiency score means a sufficient rationale.  
Similarly, in the STS task with RoBERTa-base model, the rationales extracted by ATT method have a higher plausibility score, as well as a higher sufficiency score.
Finally, we believe that other metrics can be proposed based on our benchmark.

\paragraph{Evaluation of Saliency Methods}
LIME, which uses a linear model to approximate a DL classification model, is model-agnostic and task-agnostic. It obtains the highest performance on token-F1 and sufficiency scores in SA and STS tasks, as the rationales extracted by it more accurately approximate the decision process of DL models.
But how to better apply LIME to more NLP tasks is very challenging and as the future work.

When comparing IG and ATT, we find ATT performs better on faithfulness and sentence-pair tasks.
In SA and MRC, IG performs better on plausibility and ATT method achieves better results on faithfulness, which is consistent with prior works \cite{jain-wallace-2019-attention, deyoung-etal-2020-eraser}.
In STS, ATT method achieves higher results both on plausibility and faithfulness than IG method. We think this is because the cross-sentence interaction attentions are more important for sentence-pair tasks.
Interestingly, on all three tasks, there is a positive correlation between MAP (faithfulness) and token-F1 (plausibility) with IG method.


\paragraph{Evaluation of Models}
While analyzing interpretability of model architectures, we mainly focus on IG and ATT methods, as LIME is model-agnostic.
We find that interpretability of model architectures vary with saliency methods and tasks.

Compared with transformer models, based on IG method, LSTM is competitive on plausibility and performs better on faithfulness in all three tasks. On the contrary, based on ATT method, transformer models outperform LSTM on plausibility and are competitive on faithfulness in STS and MRC tasks. As discussed above, the interaction between inputs is more important in these two tasks.

From the comparison between two transformer models with different parameter sizes, i.e., RoBERTa-base and RoBERTa-large, we find that RoBERTa-base outperforms RoBERTa-large on plausibility with these two saliency methods. Interestingly, for faithfulness evaluation, RoBERTa-base performs better than RoBERTa-large with IG method, and RoBERTa-large performs better than RoBERTa-base with ATT method.

We believe these findings are helpful to the future work on interpretability.

\section{Limitation Discussion}

We provide a new interpretability evaluation benchmark which contains three tasks with both English and Chinese annotated data. There are three limitations in our work.
\begin{itemize}[leftmargin=*]
\item How to evaluate the quality of human-annotated rationales is still open. We have several annotators to perform quality control based on human intuitions and experiences. Meanwhile, we compare model behaviors on full inputs and human-annotated rationales to evaluate the sufficiency and comprehensiveness of rationales, as shown in Table \ref{tab:rationale_eval} and Table \ref{tab:eval_rationale_more}. However, this manner has damaged the original input distribution and brings uncontrollable factors on model behaviors. Therefore, how to automatically and effectively evaluate the quality of human-annotated rationales should be studied in the future.
\item We find that the interpretability of model architectures and saliency methods vary with tasks, especially with the input form of the task. Thus our benchmark should contain more datasets of each task type ( e.g., single-sentence task, sentence-pair similarity task and sentence-pair inference task) to further verify these findings. And we will build evaluation datasets for more tasks in the future. 
\item Due to space limitation, there is no analysis of the relationships between metrics, e.g., the relationship between plausibility and accuracy, and the relationship between faithfulness and robustness. We will take these analyses in our future work.
\end{itemize}

Finally, we hope more evaluation metrics and analyses are proposed based on our benchmark. And we hope our benchmark can facilitate the research progress of interpertability.

\section{Conclusion}
We propose a new fine-grained interpretability evaluation benchmark, containing token-level rationales, a new evaluation metric and corresponding perturbed examples for three typical NLP tasks, i.e., sentiment analysis, textual similarity and machine reading comprehension.
The rationales in this benchmark meet primary properties that a rationale should satisfy, i.e., sufficiency, compactness and comprehensiveness.
The experimental results on three models and three saliency methods prove that our benchmark can be used to evaluate interpretability of both models and saliency methods. 
We will release this benchmark and hope it can facilitate progress on several directions, such as better interpretability evaluation metrics and causal analysis of NLP models.

\section*{Acknowledgements}
We are very grateful to our anonymous reviewers for their helpful feedback on this work. This work is supported by the National Key Research and Development Project of China (No.2018AAA0101900).

\bibliography{anthology,custom}
\bibliographystyle{acl_natbib}

\clearpage
\appendix

\begin{table*}[tb]
\renewcommand\tabcolsep{2.5pt}
\small
\centering
\scalebox{0.93} {
\begin{tabular}{l | c c c c c | c c c c c | c c}
\toprule
\multirow{2}{*}{Models} & \multicolumn{5}{c|}{SA} & \multicolumn{5}{c|}{STS} & \multicolumn{2}{c}{MRC}\\
\cline{2-6} \cline{7-11} \cline{12-13}
 & Acc$^f$ & Acc$^r$ & Acc$^{nr}$ & Suf & Com & Acc$^f$ & Acc$^r$ & Acc$^{nr}$ & Suf & Com & F1$^f$ & F1$^r$\\
\hline
\multicolumn{13}{c}{\textbf{English} } \\
\cline{1-13}
LSTM & 78.2 & \textbf{86.2} & 60.7 & 0.151 & 0.217 & 74.6 & 69.8 & 61.3 & 0.152 & 0.291 & 54.4 & 53.4 \\
RoBERTa-base & 93.8 & 92.4 & 70.6 & 0.084 & 0.251 & 92.7 & 89.3 & 54.8 & 0.075 & 0.418 & 71.7 & \textbf{80.8} \\
RoBERTa-large & 95.4 & 91.5 & 74.4 & 0.086 & 0.234 & 93.2 & 88.8 & 53.9 & 0.085 & 0.420 & 76.0 & \textbf{76.7} \\
\hline
\multicolumn{13}{c}{\textbf{Chinese} } \\
\cline{1-13}
LSTM & 60.0 & \textbf{70.4} & 48.7 & 0.172 & 0.135 & 75.2 & \textbf{80.7} & 51.2 & 0.083 & 0.339 & 66.4 & \textbf{82.2} \\
RoBERTa-base & 59.8 & \textbf{77.0} & 50.2 & 0.252 & 0.207 & 85.5 & \textbf{88.1} & 48.8 & 0.048 & 0.399 & 65.8 & \textbf{89.3} \\
RoBERTa-large & 62.6 & \textbf{80.6} & 47.6 & 0.212 & 0.147 & 86.0 & \textbf{87.4} & 48.9 & 0.051 & 0.433 & 67.8 & \textbf{83.3} \\
\bottomrule
\end{tabular}
}
\caption{Model performance on the original full input (Acc$^f$), human-annotated rationale (Acc$^r$), and non-rationale (Acc$^{nr}$) by removing human-annotated rationale from the original full input. $\texttt{Suf}$ and $\texttt{Com}$ represent the sufficiency score and comprehensiveness score of the human-annotated rationales, as shown in Equation \ref{equation:eraser-faithful}. We do not report F1$^{nr}$ on the MRC task, as the golden answer is not from the non-rationale.}
\label{tab:eval_rationale_more}
\end{table*}

\section{Other Details of Our Datasets}
\label{sec:data_other}

\paragraph{Other Annotation Details} We give more details about data collection, annotator information, annotation training and payment, and instructions for data usage.

\textbf{Data collection}. Except for Chinese data of SA, the annotated instances for other datasets are collected from the existing datasets, as described in Section \ref{ssec:data-collection}. In the process of collection, we ask annotators to discard instances that contain: 1) offensive content, 2) information that names or uniquely identifies individual people, 3) discussions about politics, guns, drug abuse, violence or pornography. 

\textbf{Annotator information}. We have two ordinary annotators for each task, and three senior annotators for all tasks. 
The ordinary annotators annotate the rationales and modify the rationales according to the scores from the senior annotators. They are college students majoring in languages. 
Our senior annotators are full-time employees, and perform quality control. Before this work, they have lots of experience in annotating data for NLP tasks.

\textbf{Annotation training and payment}. Before real annotation, we train all annotators for several times so that they understand the specific task, rationale criteria, etc. During real annotation, we have also held several meetings to discuss common mistakes and settle disputes. 
Our annotation project for each task lasts for about 1.5 month. And we cost about 15.5 RMB for the annotation of each instance.

\textbf{Instructions of data annotation and usage}.
Before annotation, we provide a full instruction to all annotators, including the responsibility for leaking data, disclaimers of any risks, and screenshots of annotation discussions.
Meanwhile, our datasets are only used for interpretability evaluation. And we will release a license with the release of our benchmark.

\paragraph{Data Analysis}
We report sufficiency and comprehensiveness scores of human-annotated rationales, as shown in Table \ref{tab:eval_rationale_more}. 
The sufficiency scores of human-annotated rationales are lower than those of rationales provided by transformer models or extracted by IG and ATT methods.
We can conclude that our human-annotated rationales are sufficient.
However, with IG and LIME methods, the comprehensiveness scores of human-annotated rationales are lower than those of rationales provided by models. As discussed before, the model performance on non-rationales is not accurate enough, as shown by Acc$^{nr}$, which achieves about 50\% on non-rationales. How to effectively evaluate the quality of human-annotated rationales should be studied in the future.
\begin{figure*}[tb]
\centering
\includegraphics[width=0.98\textwidth]{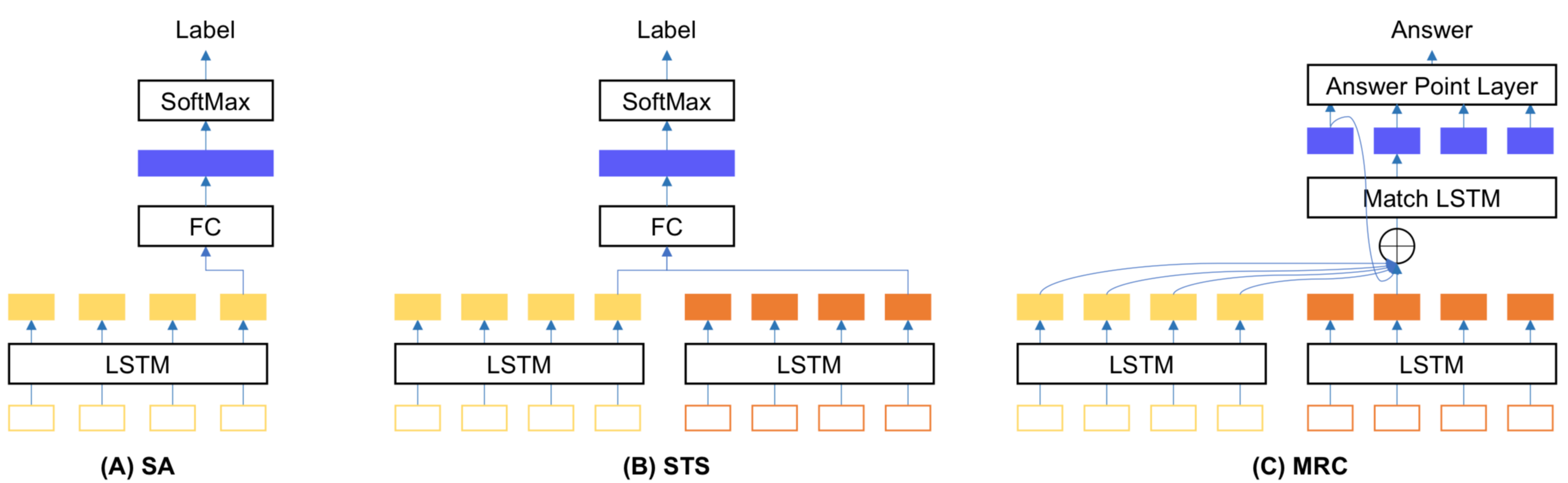}
\caption{LSTM model architectures for three tasks.}
\label{fig:lstm_models}
\end{figure*}

\begin{table*}[tb]
\renewcommand\tabcolsep{2.5pt}
\small
\centering
\scalebox{0.83} {
\begin{tabular}{l | c c c c c | c c c c c | c c c}
\toprule
\multirow{3}{*}{Models + Methods} & \multicolumn{5}{c|}{SA} & \multicolumn{5}{c|}{STS} & \multicolumn{3}{c}{MRC}\\
\cline{2-6} \cline{7-11} \cline{12-14}
 & \multicolumn{2}{c|}{Plausibility} & \multicolumn{3}{c|}{Faithfulness} & \multicolumn{2}{c|}{Plausibility} & \multicolumn{3}{c|}{Faithfulness} &
 \multicolumn{2}{c|}{Plausibility} & 
 \multicolumn{1}{c}{Faithfulness} \\
\cline{2-3} \cline{4-6} \cline{7-8} \cline{9-11} \cline{12-13} \cline{14-14} 
 & Token-F1$\uparrow$ & IOU-F1$\uparrow$ & MAP$\uparrow$ & Suf$\downarrow$ & Com$\uparrow$ & Token-F1 & IOU-F1 & MAP & Suf & Com & Token-F1 & IOU-F1 & MAP \\
\hline
LSTM + IG & 38.2 & 9.8 & 60.6 & -0.131 & 0.707 & 68.2 & 61.5 & 58.6 & 0.336 & 0.419 & 19.9 & 0.6 & 87.1\\
RoBERTa-base + IG & 35.2 & 12.5 & 51.5 & 0.118 & 0.489 & 71.9 & 71.4 & 62.1 & 0.139 & 0.470 & 34.0 & 9.1 & 67.9 \\
RoBERTa-large + IG & 37.9 & 12.9 & 43.6 & 0.123 & 0.381 & 71.8 & 72.0 & 58.1 & 0.251 & 0.547 & 25.2 & 1.7 & 61.9 \\
\hline
LSTM + ATT & 24.0 & 9.8 & 72.6 & 0.171 & 0.225 & 72.7 & 72.1 & 77.3 & 0.110 & 0.359 & 2.7 & 0.0 & 79.6 \\
RoBERTa-base + ATT & 25.7 & 6.0 & 69.5 & 0.191 & 0.320 & 67.2 & 55.4 & 71.3 & 0.201 & 0.399 & 28.5 & 5.3 & 61.4 \\
RoBERTa-large + ATT & 30.7 & 8.2 & 67.9 & 0.173 & 0.248 & 68.0 & 59.8 & 67.0 & 0.251 & 0.547 & 28.5 & 5.5 & 48.8 \\
\hline
LSTM + LIME & 38.6 & 10.1 & 59.4 & -0.130 & 0.701 & 74.8 & 79.0 & 65.9 & -0.015 & 0.411 & - & - & - \\
RoBERTa-base + LIME & 37.3 & 14.3 & 56.6 & 0.051 & 0.660 & 77.3 & 83.2 & 74.8 & -0.041 & 0.494 & - & - & - \\
RoBERTa-large + LIME & 39.0 & 14.5 & 53.0 & -0.013 & 0.653 & 76.8 & 82.9 & 74.3 & -0.024 & 0.562 & - & - & - \\
\bottomrule
\end{tabular}
}
\caption{Interpretability evaluation results on Chinese datasets of three tasks.} 
\label{tab:eval_inter_chinese}
\end{table*}

\section{Implementations Details}
\label{sec:att-extraction}

\subsection{Implementations of Evaluated Models}
\label{ssec:detail_models}
We utilize HuggingFace's Transformer \cite{wolf2019huggingface} to implement RoBERTa based models for three tasks. Please refer to their source codes\footnote{\url{https://huggingface.co/transformers/}} for more details. 
The LSTM model architectures for three tasks are shown in Figure \ref{fig:lstm_models}.

\subsection{Implementations of Saliency Methods}
\label{ssec:detail_methods}
We first describe experimental setups for three saliency methods. Then we introduce implementation details of attention-based method. Finally, we illustrate the limitations of LIME in STS and MRC tasks.

\textbf{Experimental setup}. In IG-based method, token importance is determined by integrating the gradient along the path from a defined baseline $x_0$ to the original input. In the experiments, a sequence of all zero embeddings is used as the baseline $x_0$. And the step size is set to $300$. 

LIME uses the token weight learned by the linear model as the token's importance score. For each original input, $N$ perturbed samples which contains $K$ tokens of it are created. Then the weighted square loss is used to optimize the selection of tokens that are useful for the model prediction. In the experiments, we set $N$ to $5,000$ and $K$ to $10$. In the STS task, an input is a pair of two instances. Each perturbed sample for an input consists of a perturbed example for one instance and the original input for the other instance.

\textbf{ATT method on LSTM models}. Figure \ref{fig:lstm_models} shows the architectures of LSTM models in three tasks. In the SA task, given the input instance $Q$, an LSTM encoder is used to get the representation for each token, denoted as $h^Q_i$. And a full connected layer (FC) is used to get the instance representation based on the last hidden representation. We use $h^{fc}$ to represent the representation after the FC layer. Then the instance representation $h^{fc}$ is fed into the softmax layer to get the predicted label. The attention weight for token $i$ in $Q$ is calculated by $ \frac{h^{fc} \cdot h^Q_i}{\sum^{|Q|}_{j=1} h^{fc} \cdot h^Q_j}$, where $|Q|$ represents the number of tokens in $Q$. 
Then the attention weight of the token is used as its importance score for the model prediction.

Similarly, in the STS task, the model architecture is mostly the same as that of SA. The main difference is that the input of STS consists of two instances, denoted as $Q$ and $P$, and the concatenation of their last hidden representations is fed into an FC layer. Then, referring to the attention weight calculation of $Q$, the attention weight for the token in $P$ is calculated by $ \frac{h^{fc} \cdot h^P_i}{\sum^{|P|}_{j=1} h^{fc} \cdot h^P_j}$, where $|P|$ represents the number of tokens in $P$. For each instance in a pair, we select top-$k^d$ important tokens as the rationale.

In the MRC task, the input also consists of two sequences: the question $Q$ and the passage $P$. We adopt the match-LSTM model \cite{wang2017machine} as our baseline model. The match-LSTM model uses two LSTMs to encode the question and passage respectively. Then it uses the standard word-by-word attention mechanism to obtain the attention weight for each token in the passage. And the final representation of each token in the passage is obtained by combining a weighted version of the question. We use $\bar{h}^P_i$ to represent the representation of $i$-th token in the passage. Then the importance score of $j$-th token is calculated by Equation \ref{equation:att}.
\begin{equation}
\small
a_j = \frac{\sum^{|Q|}_{i=1}e_{ij}}{|Q|} \\
\qquad e_{ij} = \frac{h^Q_i\cdot \bar{h}^P_j}{\sum^{|Q|}_{k=1} h^Q_i \cdot \bar{h}^P_k}
\label{equation:att}
\end{equation}
where $a_j$ is used as the importance score of token $j$.

\textbf{ATT method on pre-trained models}. Following related studies \cite{jain-wallace-2019-attention, deyoung-etal-2020-eraser}, on transformer-based pre-trained models, attention scores are taken as the self-attention weights induced from the [CLS] token index to all other indices in the last layer. As the pre-trained model uses wordpiece tokenization, we sum the self-attention weights assigned to its constituent pieces to compute a token's score. Meanwhile, as the pre-trained model has multi-heads, we average scores over heads to derive a final score. In the MRC task, for each token in the passage, importance score is taken as the average self-attention weights induced from this token index to all indices of the question in the last layer. 

\textbf{Limitations of LIME}. Given an input, LIME constructs a token vocabulary for it and aims to assign an important score for each token in this vocabulary. That is to say, for the token that appears multiple times, LIME neglects its position information and only assigns one score for it. However, in STS and MRC, the position of a token is very important. Therefore, It can not guarantee the effectiveness of evaluation on these two tasks with LIME.
In addition, as LIME is designed for classification models, it is difficult to apply it to the MRC task.

\section{Interpretability Evaluation on Chinese Datasets}
\label{sec:chinese_results}

We report interpretability results of three baseline models with three saliency methods on Chinese evaluation datasets in Table \ref{tab:eval_inter_chinese}. It can be seen that interpretability results on Chinese datasets have the similar trends as those on English datasets. Different from the conclusions on English datasets, on all three tasks, IG-based method outperforms ATT-based method on plausibility. And ATT method performs better than IG on faithfulness in SA and STS tasks.

\end{document}